\documentclass[lettersize,journal]{IEEEtran}
\usepackage{amsmath,amsfonts}
\usepackage{algorithmic}
\usepackage{algorithm}
\usepackage{array}
\usepackage[caption=false,font=normalsize,labelfont=sf,textfont=sf]{subfig}
\usepackage{textcomp}
\usepackage{stfloats}
\usepackage{url}
\usepackage{verbatim}
\usepackage{graphicx}
\usepackage{pifont}
\usepackage{cite}
\usepackage{lipsum}
\usepackage{amssymb,stmaryrd}
\usepackage{booktabs}

\newcommand{\bfa}{\mathbf{a}}
\newcommand{\bfb}{\mathbf{b}}

\newcommand{\bfe}{\mathbf{e}}
\newcommand{\bff}{\mathbf{f}}
\newcommand{\bfg}{\mathbf{g}}
\newcommand{\bfh}{\mathbf{h}}

\newcommand{\bfn}{\mathbf{n}}

\newcommand{\bfp}{\mathbf{p}}
\newcommand{\bfq}{\mathbf{q}}

\newcommand{\bfu}{\mathbf{u}}
\newcommand{\bfv}{\mathbf{v}}
\newcommand{\bfw}{\mathbf{w}}
\newcommand{\bfx}{\mathbf{x}}
\newcommand{\bfy}{\mathbf{y}}
\newcommand{\bfz}{\mathbf{z}}

\newcommand{\bfB}{\mathbf{B}}

\newcommand{\bfF}{\mathbf{F}}

\newcommand{\bfH}{\mathbf{H}}
\newcommand{\bfI}{\mathbf{I}}
\newcommand{\bfJ}{\mathbf{J}}
\newcommand{\bfK}{\mathbf{K}}

\newcommand{\bfP}{\mathbf{P}}
\newcommand{\bfQ}{\mathbf{Q}}
\newcommand{\bfR}{\mathbf{R}}

\newcommand{\bfT}{\mathbf{T}}

\newcommand{\bfV}{\mathbf{V}}

\newcommand{\calE}{\mathcal{E}}

\newcommand{\calM}{\mathcal{M}}

\newcommand{\calP}{\mathcal{P}}

\newcommand{\calR}{\mathcal{R}}

\newcommand{\calW}{\mathcal{W}}
\newcommand{\calX}{\mathcal{X}}
\newcommand{\calY}{\mathcal{Y}}

\newcommand{\bbR}{\mathbb{R}} 

\newcommand{\deltax}{\delta\bfx}

\newcommand{\Deltat}{\, \Delta t}

\newcommand{\bfzero}{\mathbf{0}}

\newcommand{\Log}{\text{Log}}
\newcommand{\Exp}{\text{Exp}}

\newcommand{\bfomega}{\boldsymbol{\omega}}

\newcommand{\op} {\oplus}
\newcommand{\om} {\ominus}

\newcommand{\dpar} [2]{\frac{\partial#1}{\partial#2}}

\newcommand{\pjac} [3]{\left.\dpar{#1}{#2}\right|_{#3}}
\newcommand{\Adj} [1]{\text{\textbf{Ad}}_{#1}}

\newcommand{\prx}{\hat{\bfx}} 
\newcommand{\upx}{\bar{\bfx}} 

\newcommand{\prP}{\hat{\bfP}} 
\newcommand{\upP}{\bar{\bfP}} 

\newcommand{\SO}{\mathrm{SO}(3)}
\newcommand{\SE}{\mathrm{SE}(3)}
\newcommand{\SEt}{\mathrm{SE}_2(3)}
\newcommand{\SEN}{\mathrm{SE}_N(3)}
\newcommand{\SGal}{\mathrm{SGal}(3)}

\newcommand{\bftheta}{\boldsymbol{\theta}}
\newcommand{\bfnu}{\boldsymbol{\nu}}
\newcommand{\bfrho}{\boldsymbol{\rho}}
\newcommand{\bftau}{\boldsymbol{\tau}}
\newcommand{\bfGamma}{\boldsymbol{\Gamma}}

\newcommand{\ttt}[1]{\texttt{#1}}

\usepackage{xcolor}

\begin{document}

\title{LIMOncello: Iterated Error-State Kalman Filter on the SGal(3) Manifold for Fast LiDAR–Inertial Odometry}

\author{
    Carlos Pérez-Ruiz, Joan Solà
    \thanks{All authors are with Mobile Robotics Laboratory, Institut de Robòtica
i Informàtica Industrial, Universitat Politècnica de Catalunya, Barcelona,
Spain \ttt{\{cperez,jsola\}@iri.upc.edu}}
}


\maketitle

\begin{abstract}

This work introduces \emph{LIMOncello}, a tightly coupled LiDAR--Inertial Odometry
system that models 6-DoF motion on the $\mathrm{SGal}(3)$ manifold within an
Iterated Error-State Kalman Filter backend. Compared to state
representations defined on $\mathrm{SO}(3)\times\mathbb{R}^6$, the use of $\mathrm{SGal}(3)$
provides a coherent and numerically stable discrete-time propagation model that
helps limit drift in low-observability conditions.

LIMOncello also includes a lightweight incremental i-Octree
mapping backend that enables faster updates and substantially lower memory usage
than incremental kd-tree style map structures, without relying on
locality-restricted search heuristics. Experiments on multiple real-world datasets
show that LIMOncello achieves competitive accuracy while improving robustness in
geometrically sparse environments. The system maintains real-time performance with
stable memory growth and is released as an extensible open-source implementation at
https://github.com/CPerezRuiz335/LIMOncello.

\end{abstract}

\section{Introduction}

Simultaneous Localization and Mapping (SLAM) is essential for mobile robotics, as it provides the accurate ego-motion and the consistent map required for safe planning, control, and higher-level perception. Historically, vision-based systems have been widely used, but their robustness is highly compromised on ambient illumination, texture-less scenes, and often indirect depth estimation. 

Light detection and ranging (LiDAR) sensors complement vision in such setups by providing long-range, illumination-invariant, and metrically accurate depth, which translates into reliable odometry and mapping across diverse environments. However, LiDAR introduces distinct challenges: \textit{i}) processing a massive, continuous stream of data in a timely and efficient manner, \textit{ii}) overcome the scarcity of reliable geometric features in structurally poor environments, and \textit{iii}) correcting the scan deformation caused by sensor movement during data capture \cite{FAST-LIO, FAST-LIO2}. 

To address these issues, LiDAR–Inertial Odometry (LIO) combines LiDAR with high-rate inertial measurements. High-rate Inertial Measurement Unit (IMU) measurements enable per-point deskewing and cover the latency between LiDAR frames, while LiDAR provides metric structure to correct the robot's pose. Most LIO systems are organized into a front-end and a back-end \cite{Faster-LIO}. The front-end maintains a history map and performs fast data association. This process is based on classical point-cloud registration, which works by finding the nearest neighbors for each incoming point. These neighbors are then used to fit a simple local model (typically a plane or a line) and form the residuals for optimization. 

The back-end is responsible for state optimization, with the two common families being graph-based smoothing and filtering. Under limited onboard processing, filtering based on the Extended Kalman Filter and its iterated error-state (IESKF) variant is often preferred \cite{IKFoM}, as exemplified by FAST-LIO1 \cite{FAST-LIO} and FAST-LIO2 works \cite{FAST-LIO2}.

A key step forward was the IKFoM \cite{IKFoM} framework, which formalized the IESKF on differentiable manifolds and packaged it into a practical toolkit. This gave a canonical discrete-time representation and clean separation between manifold operations and system-specific models. 

With a robust and reusable back-end in place, the community has focused effort on the front-end problems of features and computation: designing faster spatial data structures and residual evaluations that sustain real-time rates without sacrificing accuracy. The community has improved classical \mbox{k-d} tree map structures with incremental variants such as the ikd-Tree \cite{ikdtree} and sparse voxel representations like iVox \cite{Faster-LIO}, enabling efficient real-time updates and queries. Building on these advances, recent methods have shifted from point-to-map registration to point-to-feature formulations, where points are aligned with local geometric primitives (\textit{e.g.}, planes or edges) rather than individual map points \cite{VoxelMap++, C3P-VoxelMap}.

Alongside discrete-time pipelines, continuous-time motion models have gained attraction. By representing the trajectory with piecewise functions and evaluating residuals at each point timestamp, these methods improve deskewing and alignment during fast motion or asynchronous sensing. Recent systems such as DLIO \cite{DLIO} adopt point-wise continuous-time motion compensation, and others like RESPLE \cite{RESPLE} show that investing in state representation and motion modeling can yield competitive accuracy while preserving real-time performance.

Most LIO frameworks rely on what is often referred to as a \textit{compound} \cite{IKFoM} or
\textit{bundle manifold} \cite{microLie} representation of the state. In these
formulations, the system’s pose, and velocity are defined on separate sub-manifolds---usually $\SO \times \bbR^6$. While this decomposition simplifies
the implementation, it also neglects certain kinematic couplings between translational and
rotational motion, leading to an approximation that may lose accuracy under highly dynamic
motion or under sustained low-observability conditions, \textit{e.g.}, near-zero acceleration that reduces IMU observability, or degenerate LiDAR environments such as long corridors.

Beyond Euclidean and compound representations, several Lie-group manifolds have been
adopted in robotics. $\SE$ is widely used for pose estimation, while $\SEt$ became
standard in Invariant EKF-based methods
\cite{iEKF}, with extensions such as $\SEN$ \cite{SI-LIO}
and $\SGal$ explored more recently in equivariant filtering \cite{SGal_pre}.  However,
their use in tightly coupled LIO-SLAM systems remains limited.

In this direction, this letter presents \textbf{LIMOncello} (\textbf{L}ocalize \textbf{I}ntensively, \textbf{M}ap
\textbf{O}nline), the first LIO-SLAM system built upon the $\SGal$ manifold within the
IKFoM framework. To the best of our knowledge, this is the first attempt to evaluate the
suitability of $\SGal$ for tightly-coupled LIO systems.

The main contributions of this work are summarized as follows:
\begin{itemize}
    \item \textbf{Novel state representation:} We integrate the $\SGal$
manifold into IKFoM's IESKF framework, enabling consistent 6-DoF motion estimation and
improved robustness under low LiDAR observability.
    
    \item \textbf{Efficient spatial data structure:} We replace the widely used ikd-Tree
with a refactored, lightweight i-Octree implementation \cite{ioctree}, achieving faster
incremental updates and lower computational overhead by avoiding costly tree rebalancing.

\item \textbf{Modular and extensible implementation:} LIMOncello is implemented using the
\texttt{manif} library \cite{manif}, resulting in a clean, modular codebase that is easily
extensible and publicly available.

\end{itemize}

\section{Preliminaries}

\subsection{Differentiable manifolds}

Informally, a \textit{manifold} $\calM$ of dimension $m$ is a set which is locally
homeomorphic to $\bbR^m$: for every element $\calX \in \mathcal{M}$ there exists 
a local neighbourhood that can be smoothly mapped to $\bbR^m$. For a differentiable
manifold, such space always exists and is chosen as the tangent space at $\calX$, 
\textit{i.e.}, $T_{\calX} \calM$.

When the manifold $\mathcal{M}$ is also a \textit{Lie group}, the tangent space at the identity, $T_{\calE}\mathcal{M}$, is provided with a multiplication operation---typically non-associative---called the \textit{Lie bracket}, which gives $T_{\calE}\mathcal{M}$ an additional algebraic structure known as the \textit{Lie algebra}, denoted $\mathfrak{m}$. This structure encapsulates the structure of the manifold. The elements of the tangent space can be interpreted as velocities or
time derivatives, as well as error and increment variables used in
estimation and control, for elements evolving on $\mathcal{M}$~\cite{microLie}.

Lie groups are naturally endowed with the exponential and logarithmic mappings, 
$\text{exp}$ and $\text{log}$, which provide smooth isomorphisms 
between the manifold and its Lie algebra. The elements $\boldsymbol{\tau}^\wedge$ of the Lie algebra have non-trivial structures (\textit{e.g.}, skew-symmetric matrices or imaginary numbers), yet they can always be written as linear combinations of the algebra’s \textit{generators} through the \textit{wedge} $(\cdot)^\wedge$ and \textit{vee} $(\cdot)^\vee$ operators, which provide convenient mapping to $\mathbb{R}^m$, where computations can be performed in a standard vector space while preserving the group’s geometry.
\begin{figure} 
    \centering 
    \includegraphics[width=0.55\linewidth]{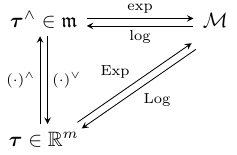} 
    \caption{Commutative diagram of the mappings between $\mathbb{R}^m$, the Lie algebra $\mathfrak{m}$, and the manifold $\mathcal{M}$, including the shortcuts $\mathrm{Exp}=\exp(\tau^\wedge)$ and $\mathrm{Log}=\log(\mathcal{X})^\vee$.}
\end{figure}

We now introduce the operators $\oplus$ and $\ominus$, which enable the combination of linear increments, expressed in the tangent vector space, with elements of the manifold.
Following the convention in \cite{microLie}, increments expressed in the local frame (\textit{e.g.}, IMU readings) are always placed on the right-hand side of the composition,%
\begin{subequations}
\begin{align}
\op &: \;\; 
\calY = \calX \op \bftau \triangleq \calX \circ \Exp(\bftau) 
\in \calM, \label{eq:rightoplus} \\[2mm]
\om &: \;\; 
\bftau = \calY \om \calX 
\triangleq \Log(\calX^{-1} \circ \calY) 
\in T_{\calX}\calM, \label{eq:rightominus}
\end{align}
\end{subequations}
where $\circ$ is the composition operation between elements of the group. 
In this letter we stick to Lie Groups that belong to $\mathrm{GL}(n)$---the space of all invertible square matrices---and the trivial Euclidean space $\bbR^n$. Then, 
$\circ$ reduces to matrix multiplication, or to vector addition, respectively, while 
$\exp$ becomes the matrix exponential for matrix Lie groups.

However, when the manifold is not a Lie group, the $\exp$ and $\log$ maps are not formally defined. In such cases, the operators $\oplus$ and $\ominus$ are usually called retraction and lift, respectively. Although their specific implementation depends on the manifold under consideration, we retain the same notation to express how a tangent perturbation modifies an on-manifold element.

For a more exhaustive introduction to Lie theory, we refer the reader to \cite{microLie}.

\subsection{IESKF on Manifolds}

In the IESKF---and ESKF---formulation, the true-state is expressed as a suitable composition (\textit{e.g.}, linear sum, matrix product) of the nominal- and the error-state. The idea is to consider the nominal-state
integrable in a non-linear fashion, and the error-state as linearly integrable and suitable 
for linear-Gaussian filtering \cite{quaternionESKF}. Ultimately, the goal is to determine
the linearized dynamics of the error-state.

The discrete dynamics of a system require a manifold $\mathcal{M}$ on which the state evolves, together with an operation that describes how the state is driven by a constant increment over a time interval $\Delta t$, assuming a zero-order hold discretization.

For Lie groups, this operation naturally corresponds to the $\oplus$ operator, since group composition directly describes how an increment in the tangent space updates the state on the manifold. For example, for $\mathbf{R}\in\mathrm{SO}(3)$ and a constant angular velocity $\boldsymbol{\omega}$ over $\Delta t$, expressed in the body frame, the update is $\mathbf{R}_{k+1} = \mathbf{R}_k\oplus(\boldsymbol{\omega} \, \Delta t)$.

When the manifold $\mathcal{M}$ is not a Lie group, the increment 
does not necessarily lie in its tangent space. Hence, we adopt the operator\footnote{The $\oplus$–$\ominus$ and $\boxplus$ notations are flipped \textit{w.r.t.} IKFoM to be consistent with the notation in Solà et al. \cite{microLie}.}
\begin{equation}
\boxplus : \mathcal{M} \times \mathbb{R}^l \to \mathcal{M}   \label{eq:boxplus}
\end{equation}

 to describe the state 
evolution. 

\paragraph{$S^2$ manifold}

In this work, the gravity vector is represented on the 2-sphere manifold $S^2(r) \triangleq \{\,\mathbf{x} \in \mathbb{R}^3 : \|\mathbf{x}\| = r,\, r > 0\,\}$, following the parameterization in \cite{IKFoM}. The $S^2$ manifold
defines a smooth surface embedded in $\mathbb{R}^3$ with constant norm $r$,
enforcing a constant-norm directional constraint. 

When $\mathbf{x}$ is represented in a fixed world frame, its state remains constant, 
$\dot{\mathbf{x}} = \mathbf{0}$, whereas in a rotating body frame it 
evolves according to $\dot{\mathbf{x}} = -\bfomega \times \mathbf{x}$, where $\bfomega$ is a constant angular velocity expressed in body frame.
Assuming a zero-order hold discretization, the discrete update in body frame becomes
\begin{equation}
 \mathbf{x}_{k+1} = {}^{\SO} \Exp(\bfomega\,\Delta t)^\top\,\mathbf{x}_k 
\;\triangleq\;
\mathbf{x}_{k+1} = \mathbf{x}_k \boxplus (\bfomega\,\Delta t).   
\end{equation}

The definitions of the $\op$ and $\om$ operators for $S^2$ are 
provided in Appendix~\ref{app:s2}.

\paragraph{Error-state linearization}

The nominal-state $\bfx$ does not take into account the noise terms $\bfw$ in its dynamics,
and other possible model nuances. As a consequence $\bfx$ accumulates error, which is collected in the error-state $\deltax$, and estimated through the IESKF \cite{quaternionESKF}. The
true-state is defined as 
\begin{equation}
    \bfx^\star = \bfx \op \deltax, \label{eq:true_state}
\end{equation}
and its dynamics are
\begin{equation}\label{eq:true_dynamics}
    \bfx^\star \leftarrow \bfx^\star \boxplus \bff(\bfx^\star, \bfu, \bfw),
\end{equation}
where $x \leftarrow f(x, \bullet)$ stands for a time update of the type $x_{k+1} = f(x_k, \bullet_k)$, $\bff$ is the function that creates a suitable increment over $\Deltat$ 
and $\bfu$ is the control signal. 

Substituting \eqref{eq:true_state} into \eqref{eq:true_dynamics}, the time evolution of the error state is defined as 
\begin{equation}
\begin{split}
\deltax_{k+1} &= \bfx^{\star}_{k+1} \om \bfx_{k+1} \\
        &\leftarrow \big( \bfx^\star \boxplus \bff(\bfx^\star, \bfu, \bfw) \big)
                    \om \big( \bfx \boxplus \bff(\bfx, \bfu, \bfzero) \big) \\
        &\leftarrow \underbrace{\big((\bfx \op \deltax) \boxplus \bff(\bfx \op \deltax, \bfu, \bfw)  \big)}_\text{A} \om \underbrace{\big( \bfx \boxplus \bff(\bfx, \bfu, \bfzero) \big)}_\text{B}
\end{split}
\end{equation}

In the IESKF prediction, the error-state is linearized as $\deltax_{k+1} \simeq \bfF_{\deltax} \deltax + \bfF_\bfw \bfw$, where
\begin{subequations}\label{eq:transition_Jacobians}
\begin{align}
  \bfF_{\deltax} &= \pjac{\deltax_{k+1}}{\deltax_k}{\deltax = \bfzero, \bfw = \bfzero}, \\
    \bfF_\bfw &= \pjac{\deltax_{k+1}}{\bfw}{\deltax = \bfzero, \bfw = \bfzero},
\end{align}
\end{subequations}
and these Jacobians can be computed by the chain rule as
\begin{subequations}
\begin{align}
    \bfF_{\deltax} &= \dpar{\text{A} \om \text{B}}{\text{A}} \left( \dpar{\text{A}}{(\bfx \op \deltax)} + \dpar{\text{A}}{\bff( \bullet)} \dpar{\bff(\bullet)}{(\bfx \op \deltax)} \right) \dpar{(\bfx \op \deltax)}{\deltax}, \label{eq:F_deltax}\\
    \bfF_\bfw &= \dpar{\text{A} \om \text{B}}{\text{A}} 
                \dpar{\text{A}}{\bff( \bullet)} \dpar{\bff(\bullet)}{\bfw} .
\end{align}
\end{subequations}

Note that when $\bfx$ lives in a Lie group, 
\begin{subequations}
\begin{align}
\bfF_{\deltax} &= {\Adj{\Exp(\bff(\bullet))}}^{-1} + \bfJ_r(\bff(\bullet)) \, \dpar{\bff(\bfx \op \deltax, \bfu, \bfw)}{(\bfx \op \deltax)}, \\
\bfF_\bfw &= \bfJ_r(\bff(\bullet)) \, \dpar{\bff(\bfx \op \deltax, \bfu, \bfw)}{\bfw},
\end{align}
\end{subequations}
where $\Adj{}$ is the Adjoint map \cite[eq.~(30)]{microLie}, and $\bfJ_r$ the right Jacobian \cite[eq.~(67)]{microLie}. 

\subsection{SGal(3)}

The Special Galilean group $\SGal$ provides a natural framework for the propagation of motion estimates over time, particularly in situations where temporal uncertainty plays a significant role \cite{mahony2025galileansymmetryrobotics}. By representing the state as a Galilean frame, time is intrinsically incorporated into the group structure, allowing spatial and temporal quantities to evolve jointly. This coupling captures the fundamental relationship between uncertainty in space and uncertainty in time, since uncertainty about \emph{when} directly translates into uncertainty about \emph{where} \cite{makingspacetimespecial}.

First used in robotics in \cite{IMUDeltas} under the name of \textit{IMU deltas matrix Lie group}, $\SGal$ is a 10-dimensional Lie group that can be interpreted as the group
of transformations that move points in both space and time---unlike $\SE$
that only represents rigid motions in 3D space.

Elements of $\SGal$ can be written as $5\times 5$ matrices,
\begin{equation}
    \boldsymbol{\Gamma} =
    \begin{bmatrix}
        \bfR & \bfv & \bfp \\
        \bfzero & 1 & t \\
        \bfzero & 0 & 1
    \end{bmatrix}
    \in \SGal \subset \bbR^{5\times 5},
\end{equation}
where $\mathbf{R} \in \mathrm{SO}(3)$ is the rotation, $\mathbf{v} \in \mathbb{R}^3$ is the linear velocity, $\mathbf{p} \in \mathbb{R}^3$ is the position, and $t \in \mathbb{R}$ is the time variable. Its Lie algebra $\frak{sgal}(3)$ is also
written as a $5\times 5$ matrix
\begin{equation}
    \bftau^\wedge = 
   \begin{bmatrix}
        [\bftheta]_\times & \bfnu & \bfrho \\
        \bfzero & 1 & \iota \\
        \bfzero & 0 & 1
    \end{bmatrix}
    \in \mathfrak{sgal}(3), \qquad
    \bftau = 
    \begin{bmatrix}
        \bfrho \\
        \bfnu \\
        \bftheta \\
        \iota
    \end{bmatrix},
\end{equation}
where $[\bftheta]_\times \in \mathfrak{so}(3)$ is the skew-symmetric matrix of 
$\bftheta \in \bbR^3$, $\bfnu \in \bbR^3$, $\bfrho \in \bbR^3$,
and $\iota \in \bbR$. 

Using $\SGal$, IMU propagation can be elegantly\footnote{By comparison, IMU propagation on $\boldsymbol{\Xi} \in \SEt$  is often expressed as 
$\boldsymbol{\Xi} \leftarrow \boldsymbol{\Xi} \op \Phi(\boldsymbol{\Xi}) \Deltat$, with $\Phi(\cdot)$ being an auxiliary mapping 
inherent to the $\SEt$ formulation, thus making the update less compact. } formulated as 
\begin{equation}
    \bfGamma \leftarrow \bfGamma \op \begin{bmatrix}
        \bfzero \\ \bfa \\ \bfomega \\ 1
    \end{bmatrix} \, \Delta t \,,
\end{equation}
assuming inputs are constant during one sampling period $\Delta t$, where $\bfomega$ and $\bfa$ are the angular rates and linear accelerations, respectively.

For the definition and derivation of the $\exp$ and $\log$ maps associated with $\SGal$, we refer the reader to \cite{kelly2025galileangroupsgal3}. 

\section{Methodology}

\subsection{System Overview}

\begin{figure*}
    \centering
    \includegraphics[width=\linewidth]{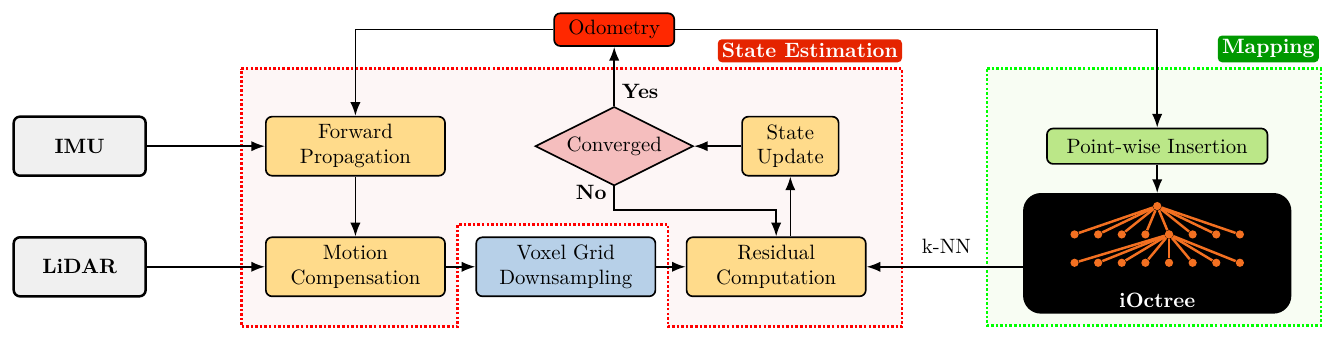}
    \caption{\textbf{System overview.} Unlike most LIO systems, motion compensation 
is done first, allowing the whole point cloud to be later used for fine-grained 
perception on a global map as detailed as possible. }
    \label{fig:pipeline}
\end{figure*}

The overall pipeline is illustrated in Fig.~\ref{fig:pipeline}. 
Incoming LiDAR scans are first deskewed using the 
IMU-integrated states, following the per-point deskewing strategy 
of \cite{FAST-LIO2}, but with the motion model formulated on $\SGal$ rather than on $\SO\times\bbR^6$. Between LiDAR 
messages, raw IMU data is continuously propagated and a window of 
intermediate poses is stored so that each LiDAR point can be 
associated with its corresponding pose during deskewing.

The downsampled point cloud is then registered against the global 
map to extract geometric correspondences, which are used to compute 
residuals for the iterated state update. Once the optimization 
converges, the current LiDAR frame is inserted into the global map, 
which is maintained as an incremental i-Octree.

\subsection{Notation}

Let the point cloud for a single LiDAR message initiated at time $t_k$
be denoted as $\calP_k$ and indexed by $k$. The point cloud $\calP_k$
is composed of points $\bfp_i \in \bbR^3$ indexed by 
$i = 1,\ldots,N$, where $N$ is the total number of points in the scan.

The world frame is denoted as $\calW$, the robot frame as $\calR$ 
(which coincides with the IMU frame $I$), and the LiDAR frame as $L$. 
The robot's state vector $\bfx$ is defined as the \textit{bundle manifold}
\begin{equation}
\begin{split}
    \bfx &= 
    \big(\, {}^\calW\bfGamma,\; {}^I\bfT_L,\; {}^\calR\bfb_{\bfomega},\; {}^\calR\bfb_{\bfa},\; {}^\calW\bfg \,\big) 
    \in \calM,\\[2mm]
    \calM &= \SGal \times \SE \times \bbR^6 \times S^2 ,
\end{split}
\end{equation}
where ${}^I\bfT_L$ are the LiDAR–IMU extrinsics, 
${}^\calR\bfb_{\bfomega}$ and ${}^\calR\bfb_{\bfa}$ are the gyroscope and accelerometer 
biases, respectively, and ${}^\calW\bfg$ is the gravity vector. With slight 
abuse of notation, frame superscripts and time indices are omitted when 
unambiguous; variables are assumed to be at time $k$ unless stated otherwise.

IMU measurements $\bfu = [\bfa,\,\bfomega]$ are modeled as
\begin{subequations}
\begin{align}
    \bfomega &= \bfomega^\star + \bfb_{\bfomega} + \bfn_{\bfomega},\\
    \bfa      &= \bfa^\star + \bfb_{\bfa} + \bfn_{\bfa} - \bfR^\top\bfg ,
\end{align}
\end{subequations}
where $\bfomega^\star$ and $\bfa^\star$ are the true angular velocity 
and linear acceleration. The terms $\bfn_{\bfomega}$ and $\bfn_{\bfa}$ 
are white measurement noise, and $\bfR$ is the system attitude. 

Biases follow a random-walk model driven by 
Gaussian noises $\bfn_{\bfb_{\bfomega}}$ and $\bfn_{\bfb_{\bfa}}$, \textit{i.e.}, $\dot{\bfb_{\bfomega}} = \bfn_{\bfb_{\bfomega}}$ and $\dot{\bfb_{\bfa}} = \bfn_{\bfb_{\bfa}}$.
All noise terms together define
\[
\bfw = 
[\, \bfn_{\bfomega},\, \bfn_{\bfa},\, \bfn_{\bfb_{\bfomega}},\, \bfn_{\bfb_{\bfa}} \, ] .
\]

\subsection{Prediction}

Based on the $\boxplus$ operation defined in \eqref{eq:boxplus}, and the transition matrices in \eqref{eq:transition_Jacobians}, the discrete propagation
of the nominal state and its covariance are
\begin{subequations}
\begin{align}
\prx &\leftarrow \prx \boxplus \bff(\prx, \bfu, \bfzero), \\
\hat{\bfP} &\leftarrow \bfF^{}_{\hat{\deltax}} \hat{\bfP} \bfF^\top_{\hat{\deltax}}
             + \bfF_{\bfw} \bfQ \bfF^\top_{\bfw}, \label{eq:propagation}
\end{align}
\end{subequations}
where $\prx$ is the propagated state, $\hat{\bfP}$ the propagated covariance, and $\bfQ$ is the covariance matrix of the noise $\bfw$ (see \cite[eq.~(262)--(265), (271)]{quaternionESKF}). Between LiDAR scans, multiple IMU measurements are integrated sequentially. At the first IMU of a given interval, $\prx$ is initialized with the last LiDAR-updated state estimate, $\upx_{k-1}$.

The function $\bff$ is given by
\begin{equation}
\bff(\bfx, \bfu, \bfw) =
\begin{bmatrix}
    \bfzero \\
    \bfa - \bfb_{\bfa} - \bfn_{\bfa} + \bfR^\top\bfg \\
    \bfw - \bfb_{\bfomega} - \bfn_{\bfomega} \\
    1 \\
    \bfzero \\
    \bfn_{\bfb_{\bfomega}} \\
    \bfn_{\bfb_{\bfa}} \\
    \bfzero
\end{bmatrix}
\Delta t \, .
\end{equation}

\subsection{Iterated Update}

The iterations within the update step are initialized using the predicted
prior, \textit{i.e.}, $\prx_j=\prx$ for $j=0$. 
At each iteration, for every downsampled point $\bfp_i$, we perform a $k$–NN search in the i-Octree map in order to fit a local plane through them, yielding the point-to-plane constraint used in \cite{FAST-LIO2}.

The measurement function is
\begin{equation}
    \bfh_i(\bfx)
    =
    \bfu_i^\top\!\left(\,\boldsymbol{\pi}(\bfGamma)\, {}^I\bfT_L\, \bfp_i - \bfq_i\right),
    \label{eq:observation_equation}
\end{equation}
where $\bfu_i$ is the normal of the plane fitted to the $k$--NN neighbors 
of $\bfp_i$, $\bfq_i$ is any point on that plane, and $\boldsymbol{\pi}(\bfGamma)$
is a smooth projection from $\SGal$ to $\SE$.

Jacobians of $\bfh_i$ \textit{w.r.t.} $\bfGamma$ 
and ${}^I\bfT_L$ are
\begin{subequations}
\begin{align}
    \dpar{\bfh_i}{\bfGamma}
    &= \bfu_i^\top
        \Big[
        \begin{array}{ccc}
            \phantom{{}^I}\bfR
            &
            \;\,-\,\bfR\!\left[\,{}^I\bfT_L\,\bfp_i\,\right]_\times
            &
            \bfzero
        \end{array}
        \Big], \\
    \dpar{\bfh_i}{{}^I\bfT_L}
    &= \bfu_i^\top
        \Big[
        \begin{array}{ccc}
            {}^I\bfR_L
            &
            -\,{}^I\bfR_L[\bfp_i]_\times
        \end{array}
        \Big], \\
    \dpar{\bfh_i}{\bfx}
    &= \Big[
        \begin{array}{ccc}
            \dpar{\bfh_i}{\bfGamma}
            &
            \dpar{\bfh_i}{{}^I\bfT_L}
            &
            \bfzero
        \end{array}
        \Big].
\end{align}
\end{subequations}

The propagated state $\prx_j$ and covariance $\prP$ from $\eqref{eq:propagation}$
impose a prior Gaussian distribution for the unknown true state $\bfx^\star_j$.
More precisely, $\prP$ represents the covariance of the following error state
\begin{equation}
\deltax_j = \bfx^\star_j \om \prx = (\prx_j \op \deltax_j) \om \prx \simeq
\prx_j \om \prx + \bfJ_j \deltax_j,
\end{equation}
where $\bfJ_j$ is the partial differentiation of $(\prx_j \, \op \, \deltax_j)\, \om \,\prx$ \emph{w.r.t.} $\deltax_j$, evaluated at $\deltax_j = \bfzero$. 

For a regular Lie group, $\bfJ_j$ coincides with the inverse of the right Jacobian \cite[eq.~(83)]{microLie}, and it transports the propagated covariance matrix $\prP$ to the tangent space at $\prx_j$, at each iteration.

Following the iterated update formulation in \cite{FAST-LIO2}, we gather all residuals into $\bfz_j$ and stack the partial derivatives of \eqref{eq:observation_equation} into $\bfH_j$, then define the projected prior covariance as
$\prP_j = (\bfJ_j)^{-1} \prP (\bfJ_j)^{-\top}$, and solve the Maximum A
Posteriori (MAP) step via
\begin{subequations}
\begin{align}
    \bfK_j
&=
\bigl( \bfH_j^\top \bfV^{-1} \bfH_j + \bfP_j^{-1} \bigr)^{-1}
\bfH_j^\top \bfV^{-1}, \\
\prx_{j+1}
    &=
    \prx_j
    \op
    \bigl(
      -\bfK_j \bfz_j
      - (\bfI - \bfK_j \bfH_j)\, \bfJ_j^{-1} (\prx_j \om \prx)
    \bigr),
    \end{align}
\end{subequations}
where $\bfV$ is the measurement covariance matrix. 

The iterated update continues until convergence,
$\|\prx_{j+1} \om \prx_j\| < \varepsilon$, or a maximum number of iterations is reached. After that the state and covariance are updated as $\upx = \prx_j$, and $\upP = (\bfI - \bfK_j \bfH_j) \prP_j$.

\subsection{i-Octree}

As in the i-Octree proposed in \cite{ioctree}, each node can have up to eight
children. This yields a shallower tree compared to a binary structure,
and even in the worst-case branching factor, the effective complexity per level
remains significantly lower. Moreover, children are indexed using Morton codes during
subdivision, speeding up traversal and query operations.

Downsampling in the ikd-Tree is based on retaining the point closest to the
voxel centroid. In contrast, downsampling in the i-Octree is performed by
recursively subdividing an octant until it contains fewer than a specified
number of points or until its spatial extent falls below a minimum threshold.

In our implementation, we omit the reallocation step that redistributes all
points inside a leaf octant during subdivision, and box-wise deletion is not
supported. Tree rebalancing, as in the original i-Octree, is also not performed,
simplifying the structure while maintaining real-time performance.

\section{Evaluation}

We evaluate LIMOncello (LIMO) on several public real-world datasets,
including \ttt{MCD} \cite{MCDDataset}, \ttt{GrandTour} \cite{GrandTour},
\ttt{R-Campus} \cite{RESPLE}, and a specific use case from the
\ttt{City Datasets} \cite{MA-LIO}. LIMO is compared against RESPLE in
its LIO configuration (R-LIO) and FAST-LIO2 (F-LIO2). All experiments
are executed on a laptop equipped with an Intel Ultra~5 125H CPU and
32\,GB RAM.

We interpolate trajectory estimates at ground truth timestamps and compute the RMSE of the absolute position error (APE) using the \textit{evo} library \cite{grupp2017evo}. Failures are denoted by \ding{55}, and the best and
second-best results are highlighted in \textbf{bold} and
\underline{underline}, respectively.

\subsection{Datasets}

\begin{figure}
    \centering
    \includegraphics[width=\linewidth]{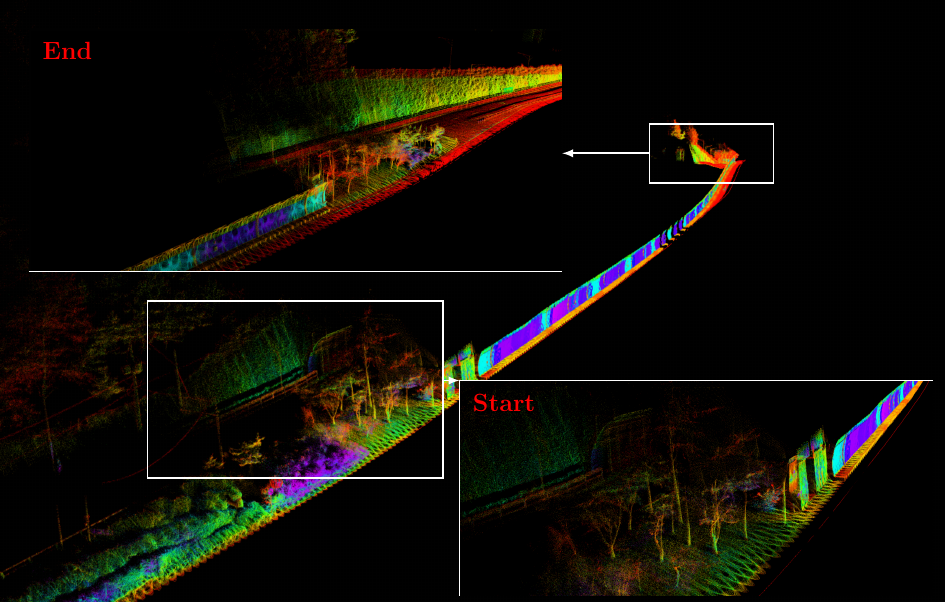}
  \vspace{0.01mm}

    \includegraphics[trim={0.cm 0cm 0cm 0cm}, clip,
                      width=\linewidth]{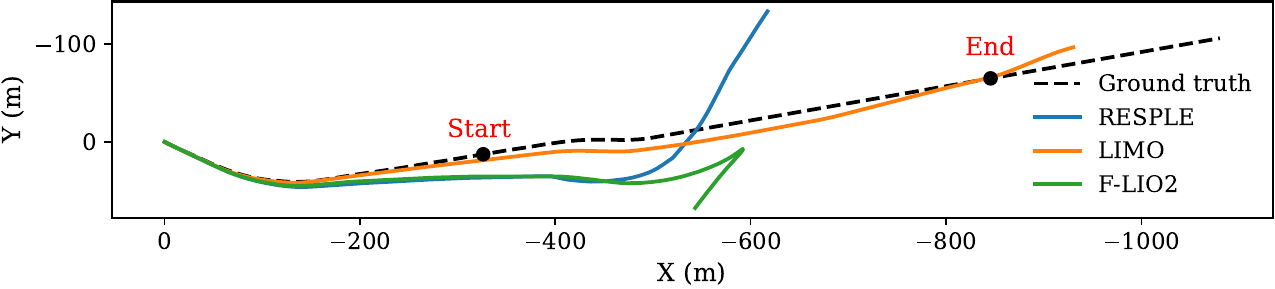}
    \caption{Qualitative comparison of accumulated trajectories in the
feature-degenerate tunnel segment of \ttt{City02}, using only the left-facing
LiDAR. \textbf{Top}: Map reconstructed by LIMO, showing both the \textit{start} and
\textit{end} points of the tunnel. \textbf{Bottom}: Trajectories projected onto the x–y plane and compared against ground truth during the tunnel traversal, where FAST-LIO2 exhibits an early,
non-recoverable divergence. RESPLE, in contrast, continues through the
tunnel but also undergoes non-recoverable drift, deviating severely from the
ground truth, including descent below the ground surface. Meanwhile, 
LIMO is the only method that traverses the tunnel under stable
conditions.}
    \label{fig:two_images}
\end{figure}

\subsubsection{\ttt{Grand-Tour}}

We evaluate indoor and outdoor sequences, including runs with mixed
stairwell passages and extended indoor–outdoor transitions. All runs
use Hesai XT32 and Novatel CPT7 sensors. LIMO demonstrates robust
performance under diverse conditions, without any failure, as shown in
Tab.~\ref{tab:grandtour}.%
\begin{table}[htbp]
    \centering
    \setlength{\tabcolsep}{4pt}
    \caption{APE (RMSE, meters) on \ttt{Grand-Tour}.}
    \begin{tabular}{@{}l|ccccc@{}}
        \toprule
        & \multicolumn{5}{c}{\textbf{Sequences}}\\
        \textbf{Method} &
        \ttt{Arc\_1} &
        \ttt{Con\_1} &
        \ttt{Con\_3} &
        \ttt{HAUS\_1} &
        \ttt{Trim\_1} \\
        \midrule
        F-LIO2 &
        $1.870$ &
        $\mathbf{0.074}$ &
        \ding{55} &
        $\underline{0.054}$ &
        $\underline{0.215}$ \\
        R-LIO &
        $\mathbf{0.088}$ &
        $\underline{0.126}$ &
        $\mathbf{0.041}$ &
        $\mathbf{0.040}$ &
        $\mathbf{0.037}$ \\
        LIMO &
        $\underline{0.371}$ &
        $0.345$ &
        $\underline{0.355}$ &
        $0.369$ &
        $0.389$ \\
        \bottomrule
    \end{tabular}
    \label{tab:grandtour}
\end{table}

\subsubsection{\ttt{MCD}}

We use day and night sequences from the NTU Campus subset of the
\ttt{MCD} dataset, recorded with Livox Mid70 and VN100 sensors.
Tab.~\ref{tab:mcd} shows that LIMO provides results comparable to
F-LIO2 and R-LIO, without encountering any failure.%
\begin{table}[htbp]
    \centering
    \setlength{\tabcolsep}{4pt}
    \caption{APE (RMSE, meters) on \ttt{MCD} (NTU Campus).}
    \begin{tabular}{@{}l|ccc|ccc@{}}
        \toprule
        & \multicolumn{3}{c}{\textbf{Day}} &
          \multicolumn{3}{c}{\textbf{Night}} \\
        \textbf{Method} & 01 & 02 & 10 & 04 & 08 & 13 \\
        \midrule
        F-LIO2 &
        $2.528$ & $0.188$ & $5.164$ &
        $\underline{0.767}$ & $2.559$ & $1.947$ \\
        R-LIO &
        $\mathbf{0.724}$ & $\underline{0.171}$ & $\mathbf{0.974}$ &
        $\mathbf{0.686}$ & $\mathbf{0.976}$ & $\mathbf{0.649}$ \\
        LIMO &
        $\underline{0.839}$ & $\mathbf{0.156}$ & $\underline{2.999}$ &
        $0.838$ & $\underline{1.375}$ & $\underline{1.636}$ \\
        \bottomrule
    \end{tabular}
    \label{tab:mcd}
\end{table}

\subsubsection{\ttt{City Datasets}}

We select a short segment from the \ttt{City02} run containing a
\(400\,\text{m}\) tunnel, used to evaluate the \(\SGal\) state
representation in a highly degenerate environment. Only the left-facing
Livox Avia is used in this scenario. LIMO is the only
method that reaches the end of the tunnel maintaining a stable trajectory, as illustrated in Fig.~\ref{fig:two_images}.

\subsubsection{\ttt{R-Campus}}

This dataset consists of a bipedal wheeled robot equipped with a Livox
Avia operating over a \(1400\,\text{m}\) campus route. The error
is evaluated using the end-to-end drift between the start and end of
each sequence. LIMO achieves an error of \(\mathbf{0.22}\,\text{m}\),
outperforming R-LIO (\(\underline{0.27}\,\text{m}\)) and F-LIO2
(\(2.70\,\text{m}\)).

\subsection{Tree Efficiency Analysis}

We benchmark the ikd--Tree,
the original i-Octree, and the proposed LIMO i-Octree under identical conditions \footnote{Both octree structures use identical parameter settings. ikd--Tree is evaluated with its default configuration.}
 on the
\ttt{ntu\_day\_01} sequence, using the OS1\textendash128 LiDAR, and
report their average processing time and memory usage. 

\begin{figure}
    \centering
    \includegraphics[width=\linewidth]{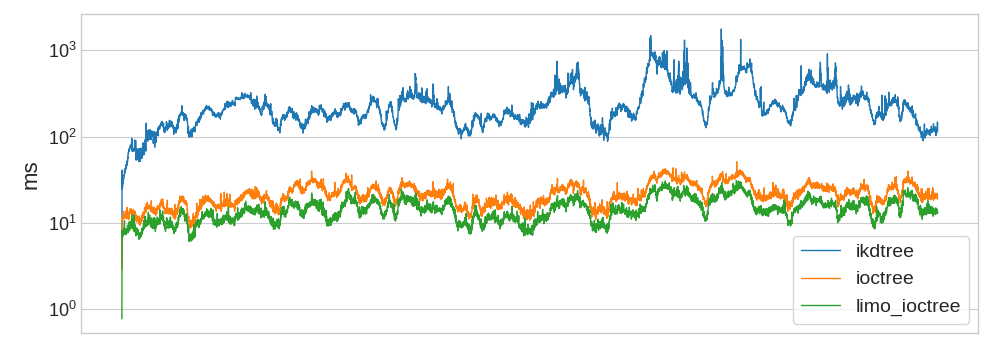}
  
    \includegraphics[trim={0.cm 0.cm 0.cm 0.cm}, clip,
                      width=\linewidth]{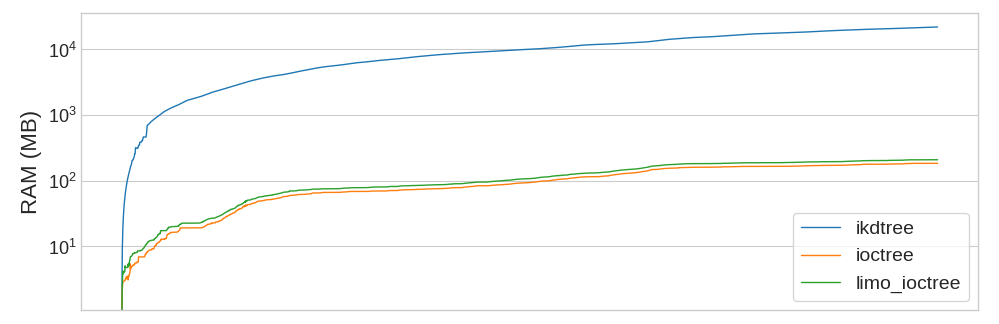}
    \caption{Per-frame computation time on the \texttt{ntu\_day\_01} sequence in a simulation setup where three IESKF iterations are executed for every LiDAR frame. Under these conditions, the octree-based methods sustain real-time performance, whereas ikd–Tree exhibits substantially higher and more variable processing times.}
    \label{fig:time_memory}
\end{figure}

As shown in
Fig.~\ref{fig:time_memory}, ikd\textendash Tree exhibits
substantially higher computational and memory overhead (avg. 248\,ms,
10.7\,GB), mainly due to its continuous rebalancing and point
redistribution. 
In contrast, both octree variants maintain real-time
performance, with average runtimes of 21.7\,ms (i-Octree) and 14.6\,ms
(LIMO i-Octree), and stable memory footprints of
approximately 100--120\,MB. Our i-Octree implementation achieves the
lowest runtime while retaining compact memory growth.

\section{Discussion}

LIMOncello achieves robust and accurate LiDAR--inertial odometry across a
diverse set of real-world datasets. Its $\SGal$-based state propagation
captures the underlying platform kinematics, resulting in consistent
motion propagation when system observability is degraded.

$\SGal$ preserves the geometric coupling between translational
and rotational dynamics, and this property is particularly beneficial in
geometrically degenerate or feature-denied environments, where several
degrees of freedom are weakly constrained by LiDAR measurements and motion
propagation dominates state evolution.

The practical impact of this design choice is most evident in the
\ttt{City02} tunnel experiment. Due to the highly degenerate geometry and
its lateral LiDAR observations, FAST-LIO2 exhibits divergence and RESPLE
experiences severe drift. LIMOncello, in contrast, maintains a stable trajectory throughout the tunnel. 

In feature-rich environments and well-conditioned trajectories, LIMOncello
achieves accuracy comparable to state-of-the-art methods such as RESPLE and
FAST-LIO2. While continuous-time splines can better model smooth motion in
these cases, LIMOncello provides a viable alternative with improved
robustness.

The proposed i-Octree mapping backend further supports practical deployment
by sustaining real-time performance, maintaining stable memory usage, and
achieving lower runtime than both the baseline i-Octree and ikd-Tree.

\section{Conclusion}

LIMOncello demonstrates that a discrete-time $\SGal$-based formulation can
deliver strong robustness in challenging operating conditions while
maintaining state-of-the-art accuracy. By providing a better
motion propagation, the proposed approach enables reliable operation in
geometrically degenerate environments.

The \ttt{City02} experiment highlights LIMOncello’s robustness in severely
feature-degenerate conditions, whereas alternative state-of-the-art approaches
diverge under the same settings. Across feature-rich environments, LIMOncello
remains comparable to existing methods while offering a simpler and more stable
state representation. Combined with a real-time i-Octree mapping backend, the
system achieves a favorable balance between robustness, accuracy, and efficiency.

\appendix
\subsection*{$S^2$ manifold, $\op$--$\om$ parameterization and Jacobians} \label{app:s2}

Following \cite{IKFoM}, the $\oplus$ and $\ominus$ operators are
defined in terms of a local tangent basis $\bfB(\bfx) \in \mathbb{R}^{3\times2}$,
whose columns span the tangent plane at $\bfx$. Specifically, for a point $\bfx \in S^2$ and a tangent increment $\bftau \in
\mathbb{R}^2$, the $\oplus$ and $\ominus$ operations are given by
\begin{subequations}
\begin{align}
\bfx \oplus \bftau &= {}^{\SO}\Exp(\bfB(\bfx)\,\bftau)\,\bfx , \\[2mm]
\bfy \ominus \bfx &= 
\bfB(\bfx)^\top 
\left(
    \theta \, \frac{\bfx \times \bfy}{\|\bfx \times \bfy\|}
\right),
\quad
\theta = \text{atan2} \!\left( \|\bfx \times \bfy\| ,\, \bfx^\top \bfy \right), \label{eq:s2-ominus}
\end{align}
\end{subequations}
where $\bfB(\bfx) = [\,\bfb_1\;\; \bfb_2\,] \in \mathbb{R}^{3\times 2}$ contains two orthonormal basis vectors spanning the tangent plane at~$\bfx$.

\begin{figure}
    \centering
    \includegraphics[width=.8\linewidth]{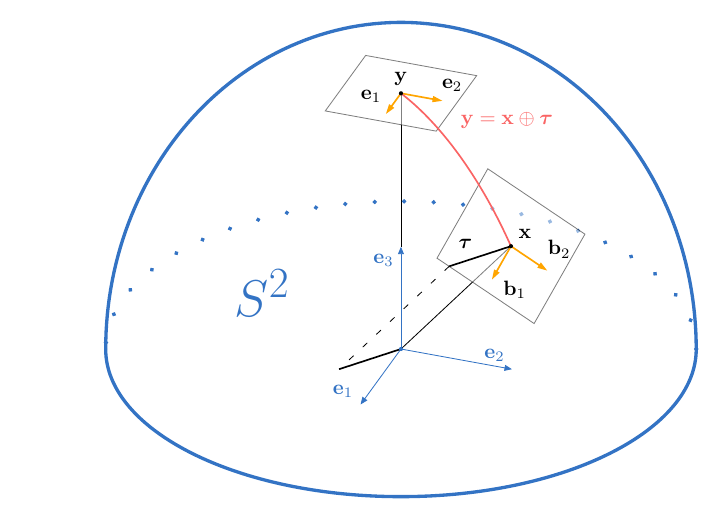}
    \caption{
Geometric interpretation of the $\oplus$ operation on $S^2$.
The tangent basis $[\bfb_1,\bfb_2]$ at $\bfx$ spans the plane
$T_\bfx S^2$, where the increment $\boldsymbol{\tau}$ is defined.
The updated point $\bfy = \bfx \oplus \boldsymbol{\tau}$ is
obtained by mapping $\boldsymbol{\tau}$ to a 3-dimensional
tangent vector via $\bfB(\bfx)$ and rotating $\bfx$ by this
vector, interpreted as an angle–axis rotation.
}
    \label{fig:S2diagram}
\end{figure}

To construct $\bfB(\bfx)$, we follow \cite{IKFoM} and define a rotation that maps the canonical basis onto a frame whose third axis aligns with $\bfx$.  
Using the reference axis $\bfe_3$, let
\begin{subequations}
\begin{align}
\bfR(\bfx) &= 
{}^{\SO}\Exp\!\left(
\frac{\bfe_3 \times \bfx}{\|\bfe_3 \times \bfx\|}
\,
\text{atan2}\!\left( \|\bfe_3 \times \bfx\| ,\, \bfe_3^\top \bfx \right)
\right), \\[2mm]
\bfB(\bfx) &= \bfR(\bfx)
\begin{bmatrix}
\bfe_1 & \bfe_2
\end{bmatrix}.
\end{align}
\end{subequations}
With this choice, increments $\bftau$ live in $\mathbb{R}^2$ as illustrated in Fig.~\ref{fig:S2diagram}.  
Closed-form Jacobians are given in~\cite[eq.~(69), (70)]{IKFoM}. 

However, when $\theta \to 0$, a first-order Taylor approximation is applied to $\text{atan2}$, and \eqref{eq:s2-ominus} simplifies to $
    \bfy \ominus \bfx 
    \simeq
    \bfB(\bfx)^\top (\bfx \times \bfy).
$
Therefore, \begin{equation}
   \frac{\partial (\bfy \ominus \bfx)}{\partial \, \bfy}
    \simeq
    \bfB(\bfx)^\top\,[\bfx]_\times .
\end{equation}

Conversely, when $\theta \to \pi$, there exist infinitely many tangent vectors that rotate $\bfx$ into $\bfy$.  Following the convention in \cite{IKFoM}, we choose
\begin{equation}
    \bfy \ominus \bfx = \begin{bmatrix} \pi \\ 0 \end{bmatrix}.
\end{equation} 




\bibliographystyle{IEEEtran}
\bibliography{references}

\end{document}